# Natural language processing of MIMIC-III clinical notes for identifying diagnosis and procedures with neural networks

**Siddhartha Nuthakki**[*]
Department of BioHealth Informatics
Indiana University-Purdue University Indianapolis
Indianapolis, IN 46202
`snuthakk@iu.edu`

**Sunil Neela**
Department of BioHealth Informatics
Indiana University-Purdue University Indianapolis
Indianapolis, IN 46202
`sneela@iupui.edu`

**Judy W. Gichoya**
Department of Radiology & Imaging Sciences
Emory University
Atlanta, GA 30322
`judywawira@emory.edu`

**Saptarshi Purkayastha**
Department of BioHealth Informatics
Indiana University-Purdue University Indianapolis
Indianapolis, IN 46202
`saptpurk@iupui.edu`

## Abstract

Coding diagnosis and procedures in medical records is a crucial process in the healthcare industry, which includes the creation of accurate billings, receiving reimbursements from payers, and creating standardized patient care records. In the United States, Billing and Insurance related activities cost around $471 billion in 2012 which constitutes about 25% of all the U.S hospital spending. In this paper, we report the performance of a natural language processing model that can map clinical notes to medical codes, and predict final diagnosis from unstructured entries of history of present illness, symptoms at the time of admission, etc. Previous studies have demonstrated that deep learning models perform better at such mapping when compared to conventional machine learning models. Therefore, we employed state-of-the-art deep learning method, ULMFiT on the largest emergency department clinical notes dataset MIMIC III which has 1.2M clinical notes to select for the top-10 and top-50 diagnosis and procedure codes. Our models were able to predict the top-10 diagnoses and procedures with 80.3% and 80.5% accuracy, whereas the top-50 ICD-9 codes of diagnosis and procedures are predicted with 70.7% and

---

[*]This is a shortened version of the Capstone Project that was accepted by the Faculty of Indiana University, in partial fulfillment of the requirements for the degree of Master of Science in Health Informatics.

63.9% accuracy. Prediction of diagnosis and procedures from unstructured clinical notes benefits human coders to save time, eliminate errors and minimize costs. With promising scores from our present model, the next step would be to deploy this on a small-scale real-world scenario and compare it with human coders as the gold standard. We believe that further research of this approach can create highly accurate predictions that can ease the workflow in a clinical setting.

# 1 Introduction

Electronic health records (EHR) consists of information about patients such as past medical and medication history, symptoms, chief complaints, treatment, procedures and tests, final diagnosis, discharge medications, and care notes or referral notes which can be tracked over time and can be considered to be the best source for evidence-based care among healthcare professionals. EHRs provides a rich source of data for clinical informatics professionals to work on. Some open resources like the Medical Information Mart for Intensive Care (MIMIC) database [1] provide a vast amount of information to work on and they are already being mined and explored in several ways [2]. MIMIC-III is an openly available extensive database developed by the Massachusetts Institute of Technology (MIT) containing deidentified healthcare data of patients admitted to critical care units of a large tertiary care hospital [3]. Data primarily exists in two forms: structured and unstructured data with diagnostics and laboratory results, medications present under structured data, whereas unstructured data includes clinician progress notes and discharge summaries. The extraction of knowledge from the structured data with the help of statistical tests and machine learning techniques is relatively easier when compared with the unstructured text. Typically, several admission notes, clinical notes, transfer notes and discharge summaries are associated with each patient's history. Extracting knowledge from this unstructured data has historically proven to be a difficult task as it requires a wide range of manual feature engineering and mapping to ontologies [4], resulting in limited adoption of such techniques.

## 1.1 Uses of NLP in healthcare

One of the recent advancements in Natural Language Processing (NLP) for healthcare is to use deep learning to find associations between unstructured text with structured data in EHR [5]. This has lead to a better understanding of the patient disease state and prognosis to improve patient outcomes [6]. In a clinical setting, NLP can be used to convert data from provider notes such as clinical notes, transfer notes, discharge notes into structured text in a predefined format which can be used for analysis. This can be an invaluable tool for Health Information Management (HIM) professionals, as they can directly process text, and organizations can utilize this to enhance communication among caregivers, provide a cost-effective way to document and process clinical text and also automate coding which is one of the important administrative tasks in documenting healthcare, billing and getting reimbursement [7, 8].Most NLP techniques have a limitation that they do not do well when dealing with non-grammatical text that has bullet points, telegraphic phrases and lack of complete sentences [9].

## 1.2 NLP work on MIMIC-III and other healthcare databases

Researchers from Massachusetts Institute of Technology developed a technique known as 'Topic modeling' to decode words in clinical notes which have multiple meanings [10], which, more recently, has been shown to be effective on MIMIC-III [11]. This technique requires minimal human supervision, automating the algorithm to refine and revise the features [12]. However, in recent years, deep learning has shown big advances in tasks like speech recognition and image processing. With the availability of open clinical data resources, deep learning in health care analytics has become a valuable resource for researchers [13, 14, 15]. A recent review paper for extracting relevant information like a diagnosis to code for medical billing, extracting symptoms, procedures relevant to the study of interest from MIMIC-III clinical notes shows dominance of deep learning approaches [16]. Single concept extraction is used for extracting vital information like diagnosis, treatments, and procedures from clinical text [17]. Many studies have applied NLP techniques to match clinical notes with the top-10 ICD codes in MIMIC-III with varied levels of success [18, 19, 20]. However, with the complexity of the clinical notes, there is significant room for improvement [21]. In this paper, we comprehensively investigate entity extraction from clinical text using a state-of-the-art NLP deep learning approach. Universal Language Model Fine-tuning (ULMFiT) [22], an effective transfer learning technique, which uses the AWD-LSTM architecture, is deployed to recognize diagnosis and procedures from clinical text.

## 2 Methodology

**Overview** We preprocessed and extracted the required data from the MIMIC-III dataset. Next, the required features were extracted according to the requirements of the language model. Lastly, we built the classifier models. Throughout this paper, we utilized Google Cloud Platform virtual machine (Tesla V100 GPU), our school's deep learning server (4x GeForce GTX 1080) and a Quadro P6000, which was through an NVidia GPU seeding grant. Figure 1 summarizes the methodology pipeline.

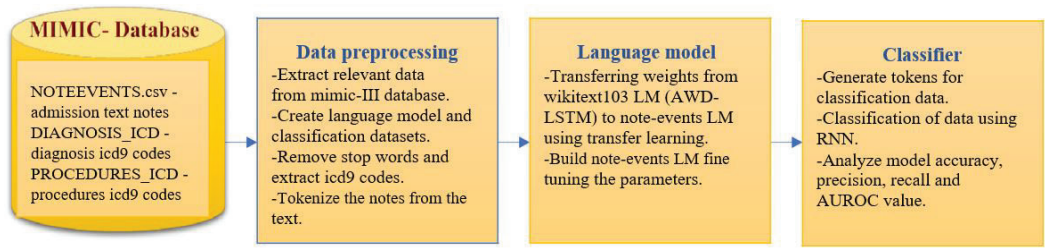

Figure 1: Methodology pipeline overview

### 2.1 Data preprocessing

We have selected the noteevents (containing the unstructured text), diagnosis_icd, procedures_icd tables from the MIMIC-III database that are relevant to the research question, as shown in Figure 2. We made use of columns such as subjectID, admissionID, discharge notes, ICD 9 codes. The selected files contained clinical notes of patients, diagnosis, procedures and ICD codes with "HADM_ID" column acting as a link for all the tables. After preprocessing the three datasets, it resulted in two separate datasets.

Initially, we merged the diagnosis and procedures datasets using the parallel data processing library, Python Dask, based on the columns, subjectID, and admissionID. After inner joining these two tables, it resulted in a new table with 3.0M rows, consisting of subjectID, admissionID, diagnosis_icd9 code and procedures_icd9 code. This table is then merged with the note-events table containing 2.0M rows which resulted in an unmanageable 800GB data which in turn was divided into multiple CSV files with each file size ranging from 8 to 12GB. We chose a file with a 10GB size from this large number of files and divided it into a train and test of 90:10. We trained the language model using this file and it took 51 hours for 1 epoch on the GeForce GTX 1080. After we realized the potential problem with the size of the data and with the constraints of time and resources, we stopped using the entire note events, instead further filtering of the diagnosis and procedures tables was performed by us.

As each subject with an admissionID has multiple diagnosis and procedures, we filtered the data by only considering the first diagnosis (most important) and procedures (first done) for each admission based on the sequence number which in turn restricted the dataset size of diagnosis and procedures to 58929 and 52243 rows respectively having 2789 and 1285 unique values [23]. Unlike merging the previous datasets wherein diagnosis was merged with procedures followed by merging with note events, herein the two datasets are individually merged with note-events data. Inner joining the diagnosis and procedures table with note-events dataset resulted in 1.8M and 1.7M rows respectively. As both the datasets are similar in size, we divided the diagnosis dataset into train and test of 90:10 ratio for training language model. The time for training 1 epoch depicted 26 hours on GeForce GTX 1080 which would take 800 to 900 hours to train a language model and classifier for 10 epochs. As allocating the resources for this long length of time is not feasible, we performed further filtering of data.

From the discussion of the literature [24], prediction of the top-10 and top-50 codes for both diagnosis and procedure tables were done. Hence, the top-10 and top-50 codes for both diagnoses and procedures were identified from the datasets containing codes only with sequence 1. Based on these codes, we filtered the entire datasets of 1.7M and 1.8M rows which were generated earlier. After the completion of the filtration process, 4 datasets (top-10 diagnosis, top-10 procedures, top-50 diagnosis, top-50 procedures - each row with clinical notes) for the final analysis remained with the top-10 and top-50 diagnosis and procedures codes and the datasets are split into 80:20 for training and testing.

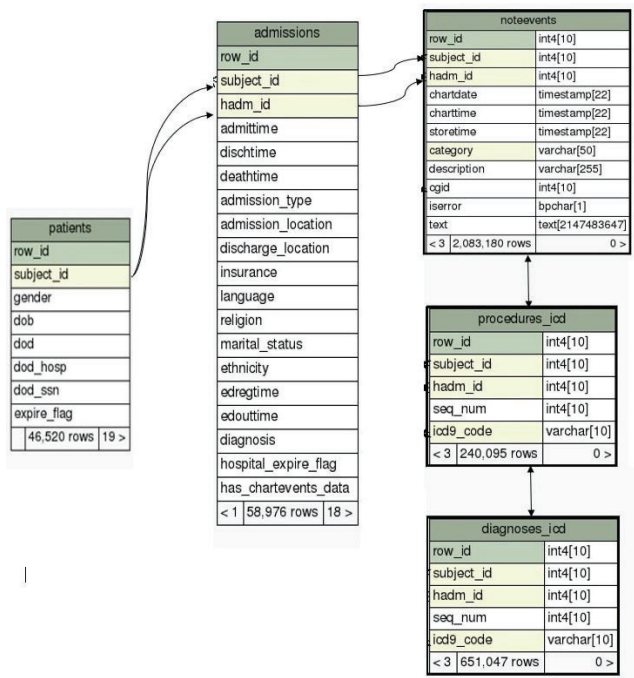

Figure 2: The schema of the tables from MIMIC

## 2.2 Feature extraction

All four datasets were processed to extract the required features. As each of the four datasets have different number of labels (10 for top 10 datasets and 50 for top 50 datasets), we converted all the labels to numeric ranging from 1 to 50 using a function called 'label encoder' in sklearn in order to be extracted and then, the datasets are divided in 80:20 train and test. As the language model requires only text without the labels, all the labels were made to zero.

'Fixup' functions are defined using Python to clean the text, remove stop words, alpha-numeric characters, and punctuation. After cleaning the text, the remaining text is tokenized using a tokenizer available in fastai which can support multi-processing that was built on top of spaCy. All the tokens are saved to the language model. We use the term frequency-inverse document frequency (TF-IDF) for feature extraction. TF-IDF which is a result of the product of TF and IDF evaluates the importance of a word in a document or a collection of documents. Term frequency can be defined as several times a word occurs in each text whereas inverse document frequency is several times a word occurs in the whole document. Words that are used more frequently are given less weight when compared to the infrequent ones. TF-IDF is obtained from tokenizing and counting the frequency of each word (TF) and finally multiplying each word with its corresponding IDF. Two TF-IDF configurations with the top 60,000 words and with a minimum frequency of 2 are used. All the tokens are converted to integers using a torch-text function called 'itos'. Along with the 60,000 tokens, two more tokens are additionally inserted, with one for unknown (_unk_) and the other for padding (_pad_). With the addition of these two tokens, total tokens accounted for goes to 60,002. A dictionary is created which can map the integer back to the string, but it doesn't cover everything as it is limited to 60,000 words. If some word which is not present in the dictionary shows up, it will return a zero automatically. This dictionary is called 'stoi' which can call for each word in the sentence and with the help of it all the tokens in the training and the test datasets are converted to numerics. These training and validation tokens of language model datasets using 'np.save' function. Likewise, all the text in the classification dataset is processed, converted to tokens, dictionary mapping of strings to integers is done and saved to the classification model folder.

### 2.2.1 Language model

The language model of our datasets is generated by a technique known as Transfer learning. We train the language model which starts with the weights from the wikitext103 language model. To load the wikitext103 weights using a torch. load, we make sure that the number of hidden and normal layers and embedding sizes are the same as that of the wikitext103 model. To map vocabulary wikitext103 to that of our dataset's vocabulary, we have used the simple 'itos' function.

A new set of weights which are zeros in vocabulary size are obtained. Later, we equate the words in the dataset with the words in the wikitext103 model. In this model, prediction of the next word is based on a set of given words and for this, a setup of a bunch of dropout values is required. We created a model data object and the model is transferred to the learner using 'learner.fit'. We first trained for a single epoch on the last layer to get the new embedding weights. Later we tested a few more epochs of the complete model. We saved the encoder and plotted the loss observed. All the language models are trained for 3 epochs each.

### 2.2.2 Classifier

Tokens for the classification data were generated similarly as that of the language model. Hyperparameters are constructed similarly but the dropouts were changed. For the classifier, we classified each text to one of the 10 or 50 classes based on the dataset. Then we passed the dataset to the data loader constructor which in turn generates a batch of that. Later, a call is made to get_rnn_classifier which creates an RNN encoder and a pooling linear classifier. We then added the number of hidden layers, dropouts and discriminated learning rates for each layer to gain more accuracy. We started training the last layer and then unfreeze another layer with much-improved accuracy being observed.

diagnosis_top10 training of 3 epochs on language model took 11.5 hours whereas the Training classifier after unfreezing took 31 hours for 10 epochs. The Procedures_top10 training of 3 epochs on the language model took 10 hours whereas the Training classifier for 10 epochs took 28 hours. In comparison, diagnosis_top50 training of 3 epochs on the language model took 14 hours whereas the Training classifier after unfreezing took 40 hours for 9 epochs. procedures_top50 training 3 epochs on language model took 17.5 hours whereas the Training classifier for 6 epochs took 33 hours. Then, we performed fine-tuning of the entire model along with training top 10 diagnosis and procedures datasets for 10 epochs and top-50 diagnoses for 9 epochs and top-50 procedures dataset for 6 epochs to receive more accurate results.

### 2.3 Using the Universal Language Model Fine-tuning for Text Classification (ULMFiT)

Recurrent neural networks (RNNs), such as long short-term memory networks (LSTMs) serve as a fundamental building block for many sequence learning tasks, including machine translation, language modeling, and question answering. To perform the text classification tasks, ULMFiT works by pretraining the language model on large-corpus and fine-tuning the single architecture [22]. In this study, we employed the ULMFiT model which uses the AWD-LSTM architecture. This architecture has dropouts at all possible layers like encoder dropout, input dropout, weight dropout, hidden dropout, and output dropout. ULMFiT technique uses three different steps in performing tasks: a) General-domain-language pretraining; b) Target task language model pretraining and c) Target task classifier fine-tuning [25]. For the suggestion mining task conducted by Haider [26], the first step included training the language model on a large corpus of Wikipedia text. The second step included fine-tuning the language model with task-specific text- the text to labeled as a suggestion or non-suggestion [26]. Finally, a classifier was added on the top of the fine-tuned language model and further fine-tuned using additional hyperparameters like dropout and unfreezing of layers [26]. In the study conducted by Dirkson and Verberne, 2019 [27], the ULMFiT model was fine-tuned on twitter health data and adverse drug effects were classified from health-related tweets. The authors (Dirkson and Verberne, 2019), fine-tuned the default language models using AWD_LSTM with a learning rate of 0.01 for the last layer and 0.001 for all layers [27]. Then a classifier is added to this fine-tuned language model with a drop out of 0.5 and a momentum of 0.8 and 0.7 [27]. For the patent classification task conducted by Hepburn, 2018 [25], ULMFiT outperformed the support vector machine (SVM) model. To achieve this text classification task the author Hepburn, 2018 [25], used the state of art language model AWD LSTM trained on Wikitext103. Then to perform the target task the language was fine-tuned which is supervised learning and then classifier is added which applies RELU activation function on the layer. In order to output the probability over the target classes, a softmax activation layer was added [25].

### 2.4 Evaluation

To check the significance of performance differences, different performance metrics like precision, recall, accuracy, f-1 score, ROC curve, and AUC value are used to evaluate different datasets, feature extraction methods, and the language and classifier, models.

## 3 Results

| S.No. | Category | No. of rows | Accuracy(%) | Precision | Recall | F-1 Score |
|---|---|---|---|---|---|---|
| 1 | Diagnosis_top10 | 677738 | 80.3 | 0.67 | 0.67 | 0.66 |
| 2 | Procedure_top10 | 632994 | 80.5 | 0.69 | 0.69 | 0.69 |
| 3 | Diagnosis_top50 | 1058928 | 70.7 | 0.58 | 0.56 | 0.55 |
| 4 | Procedures_top50 | 1215197 | 63.9 | 0.50 | 0.50 | 0.48 |

Table 1: Performance of different models

From the table, it demonstrates the performance of each model to their datasets like the top-10 and top-50 ICD codes for diagnosis and procedures respectively. The performance of the top-10 data is better with accuracy above 80.3% and 80.5% when compared to performances of the top-50 data having an accuracy of 70.7% and 63.0%. The precision and recall scores for the top-10 are also relatively better in comparison to the top-50 data.

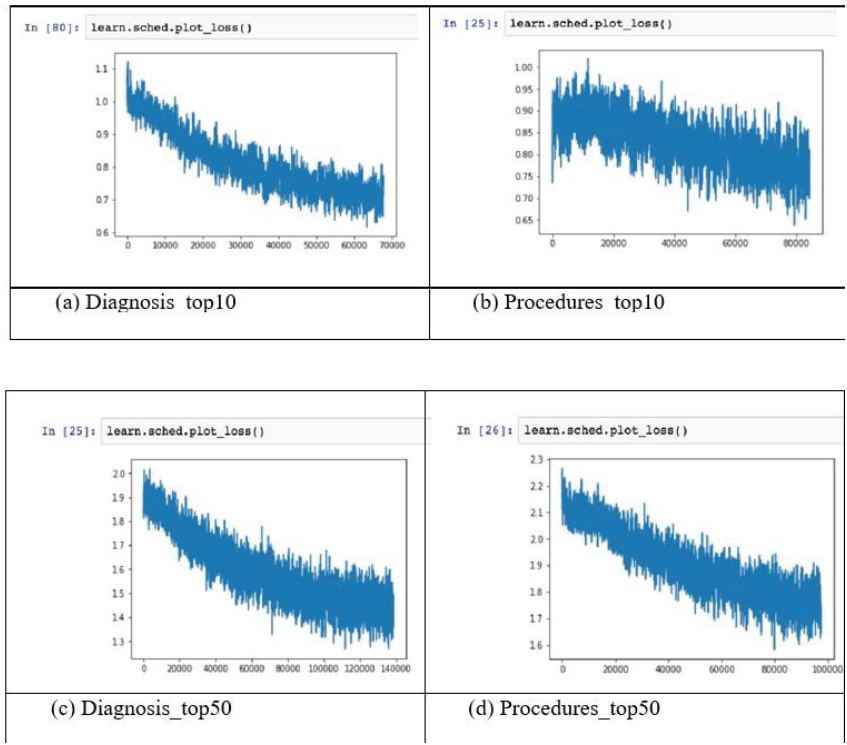

(a) Diagnosis_top10  (b) Procedures_top10

(c) Diagnosis_top50  (d) Procedures_top50

Figure 3: Plotting loss after each epoch

From Figure 3, we can infer that the loss keeps reducing with each epoch. Although the magnitude of loss is high in all the datasets, the loss is less in the top 10 diagnosis and procedures datasets when compared against its top 50 counterparts.

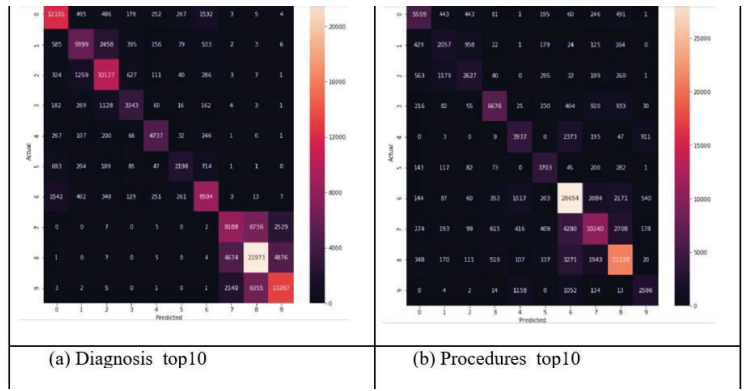

Figure 4: Confusion matrix

Figure 4 shows the values that are predicted to their actual label. From diagnosis top 10 we could observe that the last 3 diagnoses deviated more towards other labels than other diagnoses. However, in procedures top 10 classifications, all the diagnosis was accurately predicted with few deviations.

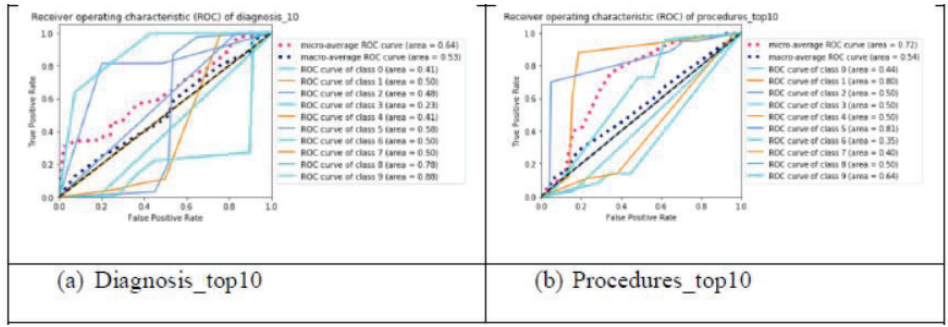

Figure 5: ROC curve

From figure 5 we can see the ROC curve and its area. The overall area for diagnosis top 10 and procedures top 10 classifiers are 64% and 72% respectively and it could be attributed to the fact that few classes were performing poorly, well under their baseline of 50%.

We could observe that the training and validation loss for diagnosis and procedures top 10 dataset is relatively less i.e., 0.3 and 0.2 respectively when compared to their counterparts which are 0.5 respectively. The closer the loss between training and validation datasets, the more accurate the model would turn out to be. Diagnosis_top10 training of 3 epochs on the language model had an accuracy of 68.1% whereas the Training classifier after unfreezing gave an impressive final accuracy of 80.3%. The Procedures_top10 training of 3 epochs on language model got an accuracy of 65.7% whereas the Training classifier for 10 epochs showed excellent accuracy of 80.5%. In comparison, Diagnosis_top50 training of 3 epochs on the language model had an accuracy of 65.5% whereas the Training classifier after unfreezing got the final accuracy of 70.7%. Procedures_top50 training 3 epochs on language model had an accuracy of 51.0% and the Training classifier for 6 epochs got an accuracy of 63.9%. We can attribute to the fact that they were performed on different GPUs with varied performances, the time taken is not directly proportional to the size of the dataset.

## 4  Discussion

Although numerous studies have reported coding of diagnosis from discharge notes, taking the entire clinical notes for each admission and predicting diagnosis and procedures is complex, but a required task for billing and reimbursement, as well as knowledge discovery in healthcare. Such automated coding of long-text clinical notes has applications in teaching medical cases, where systematic reviews of related diagnosis and procedures can be presented to medical students [28]. Moreover, these models, if with improved accuracy can help learning health systems to identify which procedures and treatments work better for certain diagnosis will be valuable. We do recognize that ICD codes are less relevant for clinical practices and ontologies like SNOMED are more descriptive, there is no reason to believe that our method will not work on codes from another dictionary or ontology. However, the goal here is to not replace human coders [29], but rather involve humans for complex tasks and for correcting such machine learning models. We observe that the ULMFiT model we have adopted for predicting the top-10 and the top-50 diagnosis and procedures do better than previous approaches and give a fairly accurate outcome. On the other hand, the accuracy of the top-50 diagnosis and procedures shows the need for better approaches if such models have to be implemented in real-world practice or even within EHR forms for automated diagnosis or decision-support systems [30]. When compared to studies conducted by Jagannatha et al. (2016) [31], who proposed Recurrent Neural Network (RNN) models for detection of medical events from datasets such as MIMIC, it could be observed that our LSTM model outperforms RNN models in achieving better prediction of codes from clinical text. From the work that carried out by Gehrmann et al. (2018) [32], it showed that deep learning methods outperformed concept extraction-based methods, which is similar to what we have discovered.

## 5 Limitations and future work

One of the challenges we faced during work is a lack of resources to run the high-end operations. Handling 800GB of data requires a huge amount of resources and time. However, the school GPU is available from the beginning of the project, as it is a shared resource, all of the GPUs were not available the whole time to train our model. Google resources were associated with high costs. Even though $6000 of credit has been given by Google, running on a large scale would have required a lot more than the available credits. The Quadro P6000, a new GPU is a grant from my mentor as a part of another project which enabled this project to move forward in this limited-resources environment. Future work relies on the prediction of diagnosis, procedures, and treatment using a single model unlike that of the multiple models predicting diagnosis and procedures distinctly. Also, some refinements are to be made to enhance the accuracy and performance of the model when more than 10 diagnoses would be employed.

## 6 Conclusion

In summary, this study observes the performance of novel variants of LSTM i.e., AWD-LSTM on MIMIC-III discharge summaries and reach to the conclusion that with this model, one can achieve good results for concept extraction i.e., diagnosis/procedure codes extraction from clinical notes. The model uses deep learning NLP techniques which assigns ICD-9 code automatically to the clinical text. Experiments demonstrate that implementation of AWD-LSTM will serve as a baseline for further research in identifying diagnosis, procedures, and treatments using a single model at once and with the automation of medical coding it can save time, cut costs that arise due to manual coding errors.